\documentclass[10pt,twocolumn,letterpaper]{article}

\usepackage{iccv}
\usepackage{times}
\usepackage{epsfig}
\usepackage{graphicx}
\usepackage{amsmath}
\usepackage{amssymb}
\usepackage{enumitem}
\usepackage{siunitx}
\usepackage{booktabs}
\usepackage{multirow}
\usepackage[pagebackref=true,breaklinks=true,letterpaper=true,colorlinks,bookmarks=false]{hyperref}

\def\ie{\text{i.e.~}}
\def\etal{\emph{et~al.}}
\def\Pluecker{Pl\"ucker~}
\def\Line{^\ell}

\usepackage{color}

\iccvfinalcopy 
\def\iccvPaperID{12412} 

\ificcvfinal\fi

\definecolor{somegray}{rgb}{0.5, 0.5, 0.5}
\newcommand{\darkgrayed}[1]{\textcolor{somegray}{#1}}
\makeatletter
\newcommand*\titleheader[1]{\gdef\@titleheader{#1}}
\AtBeginDocument{%
  \let\st@red@title\@title
  \def\@title{%
    \vskip-5em
    \bgroup\normalfont\large\centering\@titleheader\par\egroup
    \vskip1.5em\st@red@title}
}
\makeatother

\titleheader{\darkgrayed{This paper has been accepted for publication at the\\
IEEE/CVF International Conference on Computer Vision (ICCV), Paris, 2023.
\copyright IEEE}}

\title{A 5-Point Minimal Solver for Event Camera Relative Motion Estimation}

\begin{document}
\author{Ling Gao$^{1}$\thanks{indicates equal contribution} \quad Hang Su$^{1}$\footnotemark[1] \quad Daniel Gehrig$^2$ \quad Marco Cannici$^2$ \quad Davide Scaramuzza$^2$ \quad Laurent Kneip$^1$\\ \\
$^1$ Mobile Perception Lab, ShanghaiTech University, China \\
$^2$ Robotics and Perception Group, University of Zurich, Switzerland
}

\maketitle

\begin{abstract}
Event-based cameras are ideal for line-based motion estimation, since they predominantly respond to edges in the scene. However, accurately determining the camera displacement based on events continues to be an open problem. This is because line feature extraction and dynamics estimation are tightly coupled when using event cameras, and no precise model is currently available for describing the complex structures generated by lines in the space-time volume of events. We solve this problem by deriving the correct non-linear parametrization of such manifolds, which we term eventails, and demonstrate its application to event-based linear motion estimation, with known rotation from an Inertial Measurement Unit. Using this parametrization, we introduce a novel minimal 5-point solver that jointly estimates line parameters and linear camera velocity projections, which can be fused into a single, averaged linear velocity when considering multiple lines. We demonstrate on both synthetic and real data that our solver generates more stable relative motion estimates than other methods while capturing more inliers than clustering based on spatio-temporal planes. In particular, our method consistently achieves a 100\% success rate in estimating linear velocity where existing closed-form solvers only achieve between 23\% and 70\%. The proposed eventails contribute to a better understanding of spatio-temporal event-generated geometries and we thus believe it will become a core building block of future event-based motion estimation algorithms. 
\end{abstract}
\noindent \textbf{Project page: \ \footnotesize\url{https://mgaoling.github.io/eventail/}}


\section{Introduction}

\begin{figure}[t]
    \centering
    \includegraphics[width=0.485\columnwidth]{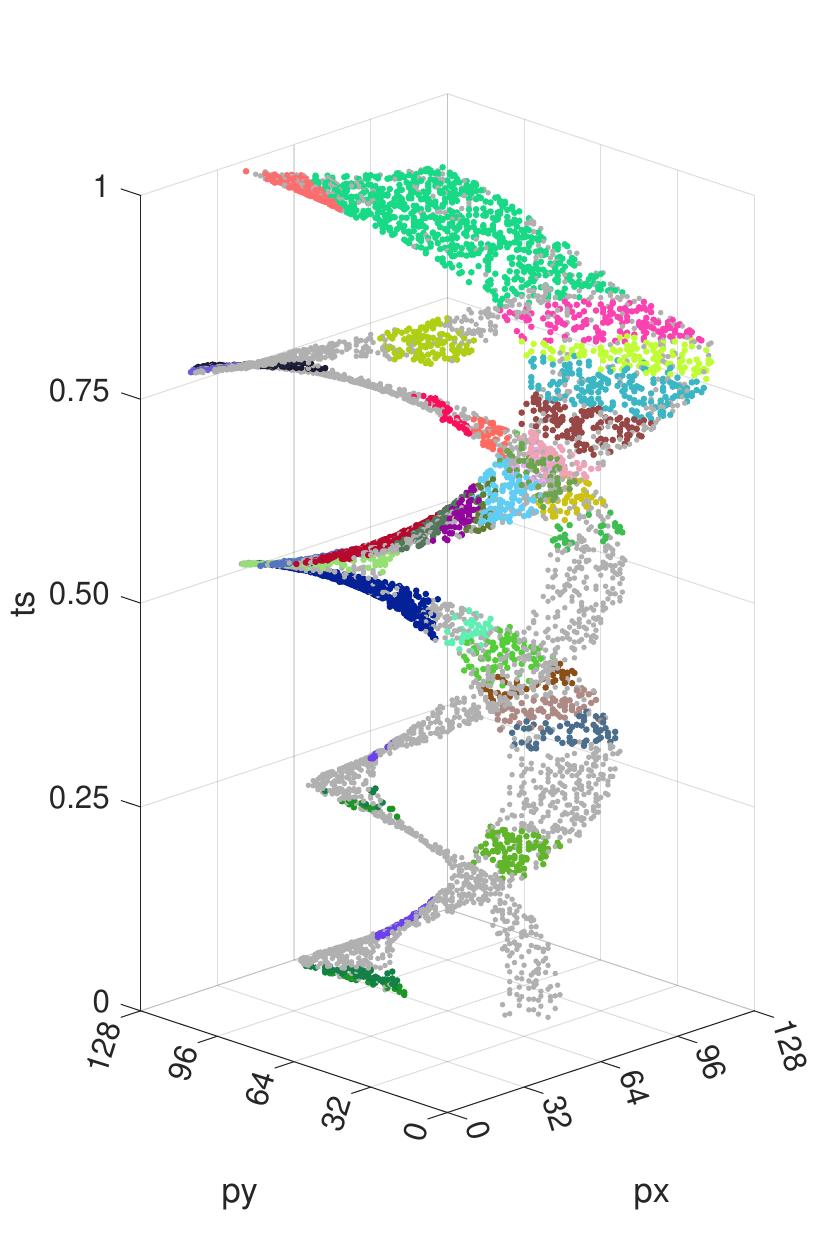}
    \includegraphics[width=0.485\columnwidth]{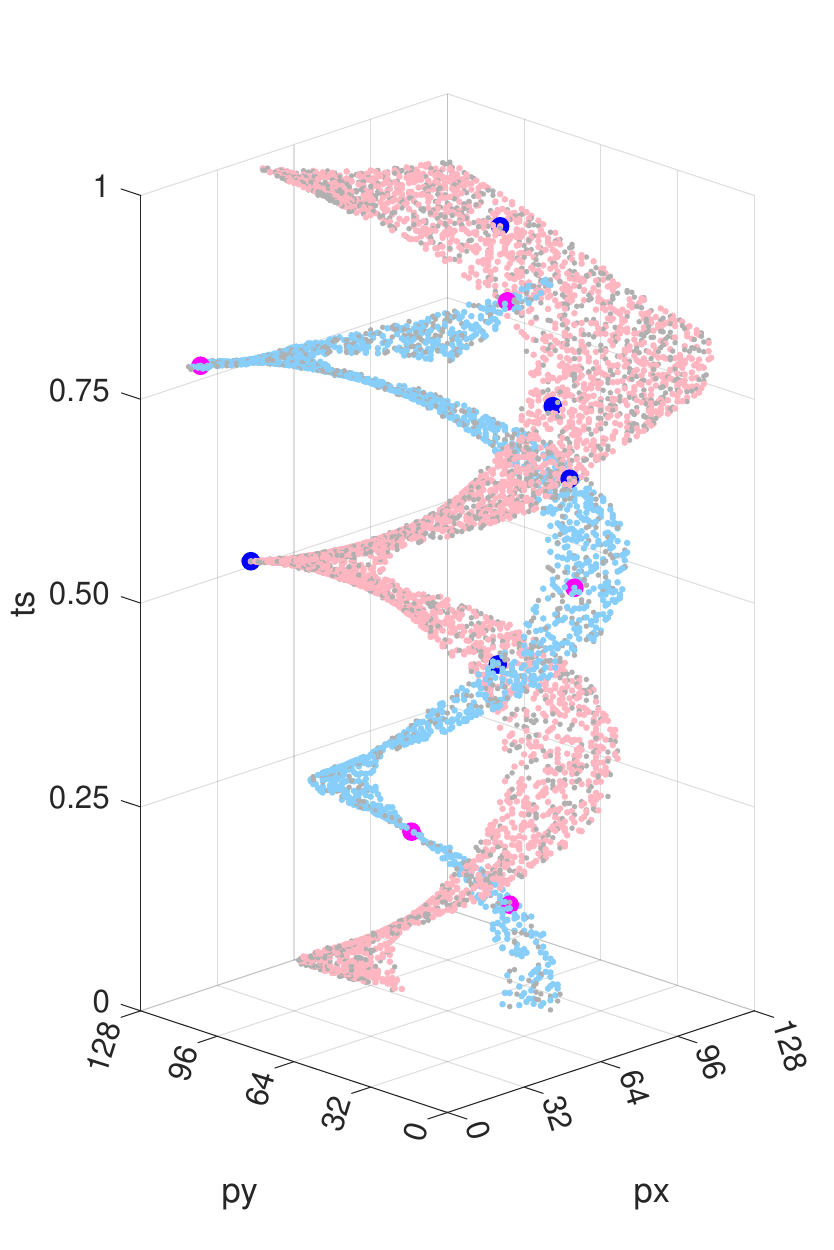}
    \caption{An event camera observing two non-parallel lines and moving with constant linear and angular velocity. The events triggered by each line lie on a manifold, which we call an eventail. We derive a minimal 5-point solver to estimate the parameters of the manifold, which includes both camera motion and scene geometry. Clustering these events based on spatio-temporal planes as done in previous work~\cite{le2020idol,peng2021continuous} generates many spurious clusters (colorful points) with many outliers (grey points). Instead, eventails result in two large clusters with fewer outliers, and a velocity direction error of only \SI{0.01}{\radian}.}
    \label{fig:front}
\end{figure}

Event-based cameras are bio-inspired vision sensors that naturally react to edges moving in the scene with microsecond temporal resolution and minimal motion blur. These intrinsic properties make events ideal for accurate relative motion estimation, especially under challenging motion and lighting conditions where standard cameras often fall short. Nevertheless, estimating motion from event measurements is an open challenge, as motion cues need to be inferred from the complex spatio-temporal structures formed by events, which typical vision-based algorithms struggle to grasp. Although event-based cameras have recently demonstrated unprecedented performance~\cite{vidal2018ultimate}, the recent development of autonomous systems has created an increased demand for more accurate and reliable solutions which could better exploit the opportunity of improved motion modeling offered by these sensors.

However, while with a traditional camera, solving for relative motion simply means aligning two views with sufficient overlap, this problem is not as straightforward to define for an event-based camera since views are not even available in the first place. Furthermore, even if the fields of view of the camera share substantial overlap at two different points in time, the structure of the perceived events at those two moments remains very much a function of local camera dynamics. In the worst case, if the camera ceases to move at all, no more events are triggered, and relative pose estimation becomes an ill-posed problem. It is intuitively clear that---for a dynamic vision sensor---the most fundamental problem of relative motion estimation is therefore given by the determination of local camera dynamics from a relatively short interval of events. The present paper introduces a geometric, deterministic solution to this problem.

The sparse and noisy nature of events has pushed the geometric vision community towards semi-dense approaches that make use of or optimize edge maps~\cite{rebecq2016evo,gallego2018unifying,zuo2022devo}. Based on the assumption that the gradient map contains straight lines, a promising area of research, therefore, looks at line features as a possible alternative to assist the geometric solution of relative event camera motion. Works in this area~\cite{le2020idol,peng2021continuous}, however, have inherent limitations that stem from a wrong assumption made during the initial feature extraction step. Indeed, they perform feature extraction independently of the relative camera displacement information, and they rely on a simple clustering strategy that models the space-time volume of events generated by a line under motion as a plane. However, as will be explained in detail in this work, lines do not form flat planes in the space-time volume of events, even if the camera undergoes constant linear and angular velocity, as evident from Figure~\ref{fig:front}.

It is thus clear that the problem of line feature extraction in the space-time volume of events can no longer be considered apart from the problem of dynamics estimation. In the present paper, we depart from this approximation and introduce a novel feature extractor that relies on a rigorously derived geometrical model of line-generated manifolds. Clustering the events of one manifold entails the identification of the manifold parameters, thereby leading to an implicit solution of the linear camera velocity from given angular rates measured by an Inertial Measurement Unit (IMU). Specifically, we make the following contributions:

\begin{itemize}[noitemsep]
\item We introduce a minimal geometric parametrization of the manifolds that contain all events generated by the observation of a single line under the assumption of locally constant, linear velocity. The parametrization involves the velocity components that are non-parallel to the line, as well as a minimal 3D parametrization of the line itself.
\item Based on this incidence relationship, we propose a minimal 5-point solver for the manifold parameters, and demonstrate its application in robust clustering for line feature detection and partial camera dynamics determination.
\item We conclude with an averaging scheme that fuses the partial camera dynamics observations from each line-generated cluster into a complete estimate of the linear camera velocity, and thereby presents a rigorous theory for deterministic event camera motion initialization.
\end{itemize}

The present paper focuses on a theoretical understanding of line-generated manifold features, and thereby contributes to a better understanding of the geometry of the temporally dense-sampling event cameras. The theory is thoroughly evaluated on simulated data, and the advantage of the method is also demonstrated in a few concluding real-world examples. In particular, we show that using our method can consistently achieve a 100\% success rate in estimating linear velocity where existing closed-form solvers achieving a success rate  23\% and 70\%. 


\section{Related work}

Vision-based camera motion estimation is a long-studied problem, and there have been countless solutions for single-camera, multi-camera, and visual-inertial scenarios. The interested reader is kindly referred to the survey of Cadena~\etal~\cite{pastpresent16} for a relatively recent overview. A work worth separate mention, though, is by Weiss~\etal~\cite{weiss134dof}, who directly estimate camera velocity.

\paragraph{Event-based motion estimation.}
The present work looks at motion estimation with an event camera, for which the last decade has already seen a number of solutions. Weikersdorfer~\etal~\cite{weikersdorfer2013simultaneous} originally propose a 2D-SLAM system with a dynamic vision sensor by employing a particle filter. The same group also proposes an event-based 3D SLAM framework by fusing events with a classical frame-based RGB-D camera~\cite{weikersdorfer2014event}. Other event-based visual odometry systems make use of known depth or 3D structure~\cite{censi2014low,mueggler2014event,gallego2016event,bryner2019event,chamorro2020high}, or are simply limited to the pure rotation scenario~\cite{gallego2018unifying}. Contrast maximization~\cite{gallego2018unifying,gallego2019focus} is proposed as a unifying framework applicable to several event-based vision tasks. It draws substantial attention~\cite{kueng2016low,stoffregen2019event1,liu2020globally,peng2020globally,peng2021globally}, but it is still limited to homographic warping scenarios. Full 6-DoF estimation is solved by Kim~\etal~\cite{kim2016real} using a filtering approach, and Rebecq~\etal~\cite{rebecq2016evo} using an alternating tracking and mapping framework. Zhu~\etal~\cite{zhu17}, Rebecq~\etal~\cite{rebecq2017real}, and Mueggler~\etal~\cite{mueggler2018continuous} furthermore propose more reliable frameworks by fusing the measurements with an IMU.

More practical 6-DoF odometry and SLAM solutions keep being proposed by fusion with other sensors. Kueng~\etal~\cite{kueng2016low} combines the event camera with a standard camera to track features and build a probabilistic map. A similar sensor combination is used in Ultimate-SLAM~\cite{vidal2018ultimate}, which improves robustness and accuracy by minimizing both vision and event-based residual errors. Zhou~\etal~\cite{zhou2021event} propose the first event-based stereo odometry system, while Zuo~\etal~\cite{zuo2022devo} use a hybrid stereo setup composed of an event and a depth camera to realize DEVO, a semi-dense edge-tracking method. Finally, a recent work by Hidalgo-Carrió~\etal~\cite{hidalgo2022event} introduces EDS, a 6-DOF monocular direct visual odometry that combines events and frames.

\paragraph{Geometry-based motion estimation.}
Event-based motion estimation can be divided into optimization-based~\cite{rebecq2016evo,le2020idol,mueggler2018continuous}, filter-based~\cite{zhu17,weikersdorfer2013simultaneous} and learning-based~\cite{maqueda2018event,gehrig2020event} solutions. However, there is a lack of research on how fundamental geometry can be applied to event-based vision. For normal cameras, Weng~\etal~\cite{weng1992motion} and Hartley~\etal~\cite{hartley1997lines} are among the first to introduce closed-form solutions for line-based motion estimation. Bartoli and Sturm~\cite{bartoli20013d,bartoli2005structure} introduce complete line-based structure from motion. More recently, based on modern line-feature extraction methods such as LSD~\cite{von2008lsd}, Zhang~\etal, Pumarola~\etal, and He~\etal~propose stereo~\cite{zhang2015building}, monocular~\cite{pumarola2017pl}, and visual-inertial fusion~\cite{he2018pl} based solutions to real-time, line feature-based SLAM, respectively.

Of particular interest to this work are event-based methods that rely on line features. Yuan~\etal~\cite{yuan2016fast} and Le~Gentil~\etal~\cite{le2020idol} present optimization-based solutions, while Peng~\etal~\cite{peng2021continuous} use tri-focal tensor geometry to present the first closed-form velocity initialization method. However, the methods make use of event-based line-feature extractors~\cite{brandli2016elised,valeiras2018event} that fail to properly parametrize the line location in both space and time. Furthermore, their minimal solver needs at least 2 lines and therefore 10 events to make a single hypothesis.

Everding and Conradt~\cite{everding18} present a low-latency line tracker, while Mitrokhin~\etal~\cite{mitrokhin20} present a learning-based method to track the surfaces generated by events. Ieng~\etal~\cite{ieng17} and Seok and Lim~\cite{seok20} finally propose model-based methods to fit and track the surfaces or curves generated by events. The perhaps most related work to ours is the work of Ieng~\etal~\cite{ieng17}, who aim at understanding the spatio-temporal sub-space properties of the surfaces generated by events. However, to the best of our knowledge, our work is the first to establish a minimal parametrization of the surface as a function of the observable 3D spatial and dynamic parameters. Furthermore, we are the first to propose a deterministic minimal solver for this problem.


\section{Theory}

We assume to have a calibrated event camera under motion observing a scene that can be approximated by a set of 3D lines. We consider a temporal slice of events from which our objective is to initialize a first-order approximation of local camera dynamics. The motion of each observed line generates its own set of events, and---while the instantaneous reprojection of a line is still a line---each event cluster is generated by a line that moves and rotates through the space-time volume of events, thereby generating a DNA-like manifold distribution. In the following, we denote such a manifold an \emph{eventail}\footnote{The word is derived from the French word \emph{\'eventail}, as the translating and rotating instantaneous line reprojection indeed resembles the geometry of the support sticks of a traditional fan.}. A set of eventails from which we wish to determine the camera velocity parameters is indicated in Figure~\ref{fig:front}.

The present section presents the theory of our method. We start with preliminaries and notations used throughout the paper. Next, we introduce a simple incidence relation that all events from one eventail need to satisfy. The constraint is transformed into minimal form, which not only reveals the intrinsic geometry of eventail manifolds and its dependency on spatio-temporal parameters, but also permits the derivation of a novel minimal solver that can be used for event clustering and a partial initialization of camera dynamics. To conclude, we present a complete velocity determination framework in which the partial observations from multiple eventails are merged into one common result.

\subsection{Notations and preliminaries}
\label{sec:notation}

We define the time interval of our slice of the space-time volume of events as $t \in [t_s - \Delta t, t_s + \Delta t]$. The set of events observed during this interval is given by $\mathcal{E}=\{\mathcal{E}_i\}_{i=1,\ldots,N}$, where $\mathcal{E}_i$ represents the manifold cluster of events corresponding to the $i$-th 3D line $\mathbf{L}_i$. Each $j$-th event of the $i$-th cluster $e_{ij}=\{u_{ij},v_{ij},t_{ij},p_{ij}\}$ is given by its $u_{ij}$ and $v_{ij}$ coordinates in the image plane, its timestamp $t_{ij}$, and its polarity $p_{ij}$\footnote{Note that, in this work, polarity is ignored.}. The camera is assumed to be calibrated and we are given a function $[u\text{ }v]^\intercal=\pi\left(\mathbf{P}\right)$ that projects points $\mathbf{P}\in\mathbb{R}^3$ defined in the camera frame into the image plane. Conversely, we are given the inverse function $\mathbf{f}=\pi^{-1}\left(u,v\right)$ that transforms image plane coordinates into a unit-norm 3D direction vector defined in the camera frame and pointing towards the corresponding point in 3D. We furthermore define the first-order dynamics parameters $\mathbf{v}$ and $\boldsymbol{\omega}$, which represent the instantaneous translational and rotational velocity of the camera, respectively\footnote{Note that the constant velocity motion assumption is violated if the camera experiences more jerky motion. One may argue that the approximation still holds within smaller time intervals, which will however limit the number of events and the baseline experienced over the course of the interval, and thus reduce the observability of depth and motion. As with any constant velocity approximation, our approach depends on sufficiently smooth motion.}.

A line $\mathbf{L}$ in $\mathbb{R}^3$ can be represented by its direction vector $\mathbf{d}$ and a point $\mathbf{P}$ that lies on the line. The \emph{\Pluecker coordinates} of this line are then defined as $[\mathbf{d}^\intercal \text{ } \mathbf{m}^\intercal]^\intercal \doteq [\mathbf{d}^\intercal \text{ } \left(\mathbf{P} \times \mathbf{d}\right)^\intercal]^\intercal$, and the vector $\mathbf{m}$ is referred to as the \emph{moment vector}. If two nonparallel lines $\mathbf{L}_1 = [\mathbf{d}_1^\intercal \text{ } \mathbf{m}_1^\intercal]^\intercal$ and $\mathbf{L}_2 = [\mathbf{d}_2^\intercal \text{ } \mathbf{m}_2^\intercal]^\intercal$ intersect, the following product is required to vanish, \ie
\begin{equation}
\label{eq:line_intersection}
    \langle \mathbf{d}_1 \,, \mathbf{m}_2 \rangle + \langle \mathbf{d}_2 \,, \mathbf{m}_1 \rangle = 0 \,,
\end{equation}
where $\langle \cdot\,, \cdot \rangle$ stands for the inner product. Note that, as long as the moment vector is strictly defined as the cross-product of a point on the line and the exact same direction vector that is used as the first three entries of the \Pluecker vector~\cite{plucker65}, the constraint is generally valid and there is no requirement on the norm of the direction vector.

\subsection{Incidence Relationship}

The entire problem is formulated relative to the camera frame at time $t_s$. Let $\mathbf{L}_i=[\mathbf{d}_i^\intercal \text{ } \mathbf{m}_{i}^\intercal]^\intercal$ be the \Pluecker coordinates of a line defined in the said reference frame. $e_{ij} \in \mathcal{E}_i$ is the $j$-th event triggered by the moving reprojection of the $i$-th line $\mathbf{L}_i$. Using our first-order dynamics approximation, we can easily define the position of the camera center at time $t_{ij}$ seen from the reference frame at time $t_s$ as $\mathbf{C}\left[t_{ij}\right] = \mathbf{v}\cdot (t_{ij}-t_s)$, and the rotation that takes points from the camera frame at time $t_{ij}$ back to the reference frame at time $t_s$ as $\mathbf{R}\left[t_{ij}\right]=\operatorname{exp}(\lfloor\boldsymbol{\omega}\rfloor_{\times}\left(t_{ij}-t_s\right))$. Here, $\lfloor \boldsymbol{\omega} \rfloor_{\times}\in \mathbb{R}^{3\times 3}$ is the skew-symmetric matrix formed from the angular rate $\boldsymbol{\omega}\in\mathbb{R}^3$. Note that, similar to~\cite{peng2021continuous}, we assume the angular velocity of the camera to be known, as it can be reliably measured by modern IMU's. We will present a sensitivity analysis to these measurements in the experimental section. Last, the direction vector pointing from the camera frame at time $t_{ij}$ to the 3D point on the line $\mathbf{L}_i$ that triggered the event is given by $\mathbf{f}_{ij}=\pi^{-1}\left(u_{ij},v_{ij}\right)$.

The measurement of the event can now easily be expressed as a ray defined in the reference frame at time $t_s$ where $\mathbf{R}\left[t_{ij}\right] \mathbf{f}_{ij}$ is the direction of the ray, and $\mathbf{C}\left[t_{ij}\right]$ represents the origin of the ray (\ie a point on the line). The ray is therefore described by the \Pluecker coordinates
\begin{equation}
\left[ \left(\mathbf{R}\left[t_{ij}\right] \mathbf{f}_{ij}\right)^\intercal \text{ } \left( \mathbf{C}\left[t_{ij}\right] \times \left(\mathbf{R}\left[t_{ij}\right] \mathbf{f}_{ij}\right) \right)^\intercal \right]^\intercal \,.
\end{equation}
Finally, the incidence relation is given by \eqref{eq:line_intersection}, namely
\begin{equation}
\label{eq:nonminimalIncidence}
\langle \mathbf{d}_{i} \,,  \mathbf{C}\left[t_{ij}\right] \times \left(\mathbf{R}\left[t_{ij}\right] \mathbf{f}_{ij}\right) \rangle +
\langle \mathbf{R}\left[t_{ij}\right] \mathbf{f}_{ij} \,, \mathbf{m}_i \rangle = 0 \,.
\end{equation}

\begin{figure}
    \centering
    \includegraphics[width=0.8\columnwidth]{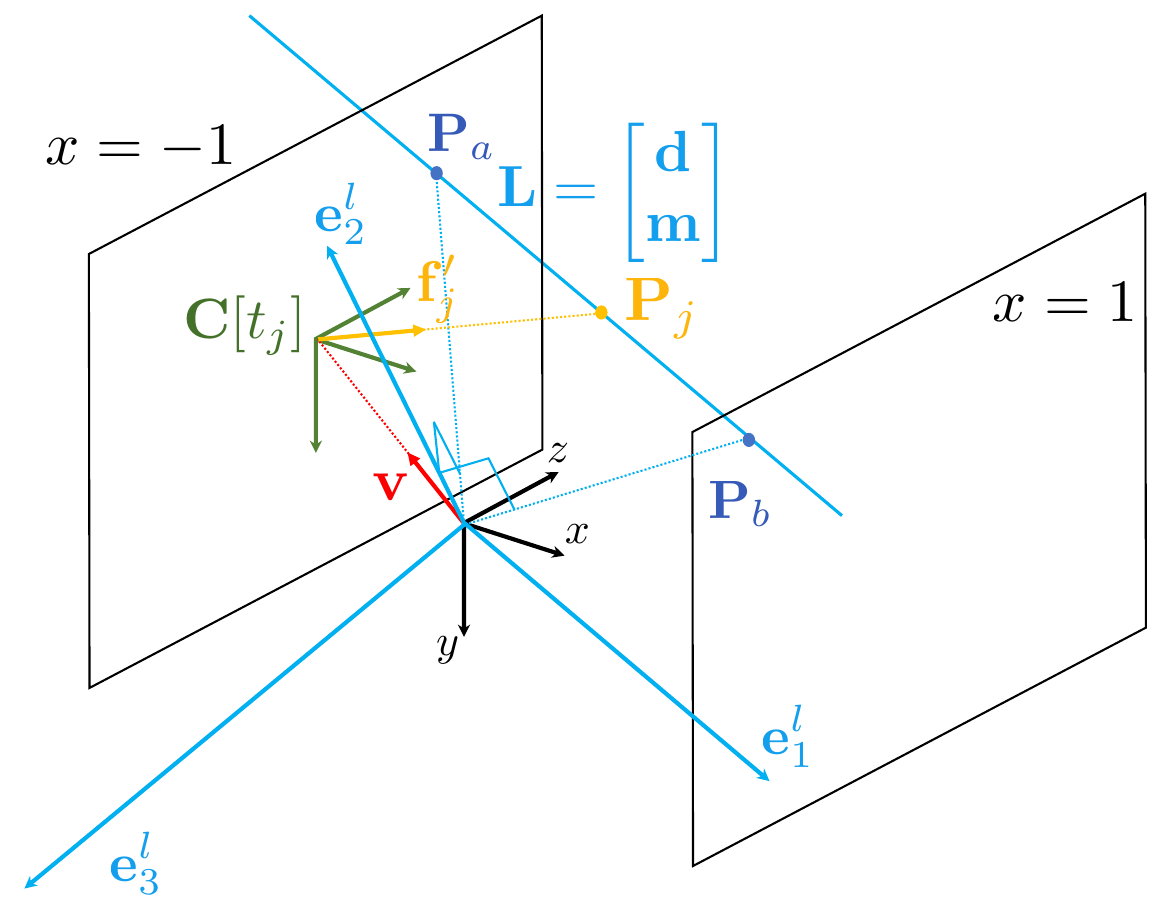}
    \caption{Incidence relationship between the line $\mathbf{L}$ with the two-point-two-plane parametrization, and the event with the bearing vector $\mathbf{f}_{j}'$. Camera velocity is given in the line-dependent reference frame $\mathbf{R}_{\ell}=[\mathbf{e}\Line_{1} \, \mathbf{e}\Line_{2} \, \mathbf{e}\Line_{3}]$.}
    \label{fig:geometry}
\end{figure}

Next, we assume that rotational velocity is already known. In practice, this assumption is easily satisfied by the addition of an IMU. In the continuation, we therefore assume that events are directly represented by their unrotated, normalized coordinates $\mathbf{f}_{ij}'=\mathbf{R}\left[t_{ij}\right] \mathbf{f}_{ij}$. Our incidence relation becomes
\begin{equation}
\label{eq:nonminimalIncidence2}
\langle \mathbf{d}_{i} \,,  \left(\mathbf{v}\cdot t_{ij}'\right) \times  \mathbf{f}_{ij}' \rangle +
\langle \mathbf{f}_{ij}' \,, \mathbf{m}_i \rangle = 0 \,,
\end{equation}
where $t_{ij}'=t_{ij}-t_s$. Figure \ref{fig:geometry} indicates the geometry of the problem. The above incidence relation is already in polynomial form, and relates our measurements $\mathbf{f}_{ij}'$ and $t_{ij}'$ to the remaining unknowns $\mathbf{v}$ and $\mathbf{L}_i$. However, it is not a minimal parametrization of the eventail manifold because \Pluecker coordinates are not minimal representations of lines, and the velocity component that is parallel to the line indeed has no influence on the incidence condition (\ie it is unobservable, as sliding the camera along the direction of the line is unable to cause residual errors, a condition which is also known as the \textit{aperture problem}).

\subsection{Transition into a minimal form}

The following three subsections are with respect to a single cluster, only, which is why index $i$ is dropped from the formulation. The transition into minimal form consists of two steps. We start by replacing the representation of the unknown 3D line by a two-point-two-plane parametrization~\cite{hartley2003multiple}, which is minimal. More specifically, we define the 3D line $\mathbf{L}$ by the intersection points with the planes $x=1$ and $x=-1$ defined in the reference frame at time $t_s$. The two points in the reference frame are given by
\begin{equation}
    \mathbf{P}_a = [-1, y_a, z_a]^\intercal \,, \ 
    \mathbf{P}_b = [ 1, y_b, z_b]^\intercal \,.
\end{equation}
As required, the new, minimal parametrization has only four degrees of freedom. Note that the orientation of the support planes for the two points can be arbitrarily changed, and their normal vectors can be chosen such that they are parallel to the approximate direction of the line in the image plane at time $t_s$ (the latter can be set by for example considering the line connecting event samples). Without loss of generality, here we assume that the reference frame is already defined such that the support planes are given as defined above.

Next, we take care of the above-discussed problem that only part of the velocity vector is observable from a single-line observation under constant first-order dynamics. A minimal parametrization of the eventail manifold can only depend on part of the velocity vector, which notably are the components that are orthogonal to the line direction vector. To explicitly parametrize the vector as such, we introduce an intermediate, line-dependent reference frame in which the observable part of the camera velocity can be simply defined. The intermediate reference frame is given by
\begin{equation}
    \mathbf{e}_1\Line = \mathbf{P}_b - \mathbf{P}_a \,, \
    \mathbf{e}_2\Line = \mathbf{P}_b \times \mathbf{P}_a \,, \
    \mathbf{e}_3\Line = \mathbf{e}_1\Line \times \mathbf{e}_2\Line \,.
\end{equation}
$\mathbf{e}_1\Line$ is parallel to the line direction, $\mathbf{e}_2\Line$ is orthogonal to plane traversing the camera center and the line $\mathbf{L}$, and $\mathbf{e}_3\Line$ is contained in the latter plane but pointing orthogonally away from the line. Only velocity components along $\mathbf{e}_2\Line$ and $\mathbf{e}_3\Line$ are observable, hence the velocity can be substituted by
\begin{equation}
  \mathbf{v} = \left[ \begin{matrix} \mathbf{e}_1\Line & \mathbf{e}_2\Line & \mathbf{e}_3\Line \end{matrix} \right] \cdot \left[ \begin{matrix} 0 & v\Line_y & v\Line_z \end{matrix} \right]^\intercal
  = \mathbf{R}_{\ell} \mathbf{v}_{\ell} \,.
\end{equation}
As required, this parametrization has only two additional degrees of freedom (\ie $v_y$ and $v_z$). Note that $\mathbf{R}_{\ell}$ is not an orthonormal rotation matrix, but it only represents an orthogonal basis for the minimal definition of the velocity. Using the new parameters, incidence relation \eqref{eq:nonminimalIncidence2} becomes
\begin{equation}
    \label{eq:minimalIncidence}
        t_{j}'(\mathbf{P}_{b} - \mathbf{P}_{a})^\intercal ((\mathbf{R}_{\ell}\mathbf{v}_{\ell})\times \mathbf{f}_{j}') 
        -{\mathbf{f}_{j}'}^\intercal (\mathbf{P}_{b} \times \mathbf{P}_{a}) = 0.
\end{equation}
The two-point-two-plane parametrization and the intermediate, line-dependent reference frame are shown in Figure~\ref{fig:geometry}.  

\subsection{Elementary properties of the minimal form}

Before introducing the solver, here we list two corollaries along with their corresponding proofs.

Corollary 1: \textit{The solution in terms of motion and structure parameters is scale invariant. In the case of our parametrization, the scale invariance is entirely reflected by the structure parameters. Scaling the line $\mathbf{L}$ will scale the velocity basis vectors $\mathbf{e}_2\Line$ and $\mathbf{e}_3\Line$ such that $v\Line_y$ and $v\Line_z$ remain unchanged. The latter are constant ratios.}

\textbf{Proof:} It is sufficient to prove that if scaling the line points $\mathbf{P}_a$ and $\mathbf{P}_b$ by a factor $k$, the basis vectors $\mathbf{e}_2\Line$ and $\mathbf{e}_3\Line$ will be equally scaled by $k$. Let us denote the intersection points with planes $x=-1$ and $x=1$ of the scaled line by $\mathbf{P}_a'$ and $\mathbf{P}_b'$. $\mathbf{P}_a'$ and $\mathbf{P}_b'$ must lie on the line connecting $k\mathbf{P}_a$ and $k\mathbf{P}_b$ such that their first coordinate equals to -1 or 1, respectively. We therefore have
\begin{eqnarray}
  & & \left\{ \begin{matrix}
  \mathbf{P}_a' = k\mathbf{P}_a'+\lambda_a (k\mathbf{P}_b - k\mathbf{P}_a) = [-1 \, \cdot \, \cdot]^\intercal \\
  \mathbf{P}_b' = k\mathbf{P}_a'+\lambda_b (k\mathbf{P}_b - k\mathbf{P}_a) = [1 \, \cdot \, \cdot]^\intercal
  \end{matrix} \right. \label{eq:scaled_points} \\
  & \Leftrightarrow & \left\{ \begin{matrix}
  2\lambda_a k - k = -1 \\
  2\lambda_b k - k = 1
  \end{matrix} \right. \Leftrightarrow \left\{ \begin{matrix}
  \lambda_a = \frac{k-1}{2k} \\
  \lambda_b = \frac{k+1}{2k}
  \end{matrix} \right. \nonumber \, .
\end{eqnarray}
Back substituting in \eqref{eq:scaled_points}, we obtain
\begin{equation}
  \left\{ \begin{matrix}
  \mathbf{P}_a' = \frac{k}{2} (\mathbf{P}_a + \mathbf{P}_b) - \frac{(\mathbf{P}_b-\mathbf{P}_a)}{2} \\
  \mathbf{P}_b' = \frac{k}{2} (\mathbf{P}_a + \mathbf{P}_b) + \frac{(\mathbf{P}_b-\mathbf{P}_a)}{2}
  \end{matrix} \right. \label{eq:scaled_points2} \, .
\end{equation}
The new basis vector $\mathbf{e'}_2\Line$ is finally given by
\footnotesize
\begin{eqnarray}
  \mathbf{e'}_2\Line & = & \mathbf{P}'_b \times \mathbf{P}'_a = -k(\mathbf{P}_a+\mathbf{P}_b)\times \frac{\mathbf{P}_b-\mathbf{P}_a}{2} \nonumber \\ & = & -k \left( \frac{\mathbf{P}_a\times\mathbf{P}_b}{2} - \frac{\mathbf{P}_b\times\mathbf{P}_a}{2} \right) = k \left( \mathbf{P}_b \times \mathbf{P}_a \right) = k \mathbf{e}_2\Line \,. \nonumber
\end{eqnarray}
\normalsize
Next, it is easy to see that $\mathbf{e'}_1\Line = \mathbf{P}'_b - \mathbf{P}'_a = \mathbf{P}_b - \mathbf{P}_a = \mathbf{e}_1\Line$. Subsequently, $\mathbf{e'}_3\Line = \mathbf{e'}_1\Line \times \mathbf{e'}_2\Line = k \mathbf{e}_1\Line \times \mathbf{e}_2\Line = k \mathbf{e}_3\Line$.\hspace{0.9cm}$\blacksquare$

Corollary 2: \textit{There exists a solution duality, \ie if $\{\mathbf{P}_{a}, \mathbf{P}_{b}, \mathbf{v}_{\ell}\}$ is a valid solution, then $\{\mathbf{P}'_{a} = -\mathbf{P}_{b}, \mathbf{P}'_{b} = -\mathbf{P}_{a}, \mathbf{v}'_{\ell} = \mathbf{v}_{\ell}\}$ is also a solution. It notably corresponds to a reversal of the velocity and a placement of the line behind the camera. It is furthermore interesting to note that only the line parameters are affected by the solution duality.}

\textbf{Proof:} $\mathbf{e'}_2\Line = \mathbf{P}'_{b}\times\mathbf{P}'_{a} = -(\mathbf{P}_{b}\times\mathbf{P}_{a})=-\mathbf{e}_2\Line$, and $\mathbf{e'}_1\Line = \mathbf{P}'_{b}-\mathbf{P}'_{a} = \mathbf{P}_{b}-\mathbf{P}_{a}=\mathbf{e}_1\Line$, which is why $\mathbf{e'}_3\Line = -\mathbf{e}_3\Line$. Let $\mathbf{R}'_{\ell} = \left[ \begin{matrix} \mathbf{e'}_1\Line & \mathbf{e'}_2\Line & \mathbf{e'}_3\Line \end{matrix} \right]$. Given that the first coordinate of $\mathbf{v}_{\ell}$ is always zero, we therefore have $\mathbf{R}'_{\ell} \cdot \mathbf{v}'_{\ell} = -\mathbf{R}_{\ell} \cdot \mathbf{v}_{\ell}$. Finally, for our incidence relation, we have
\begin{eqnarray}
  & & t'_{j}(\mathbf{P}'_{b} - \mathbf{P}'_{a})^\intercal ((\mathbf{R}'_{\ell}\mathbf{v}'_{\ell})\times \mathbf{f}'_{j}) 
        -{\mathbf{f}_{j}'}^\intercal (\mathbf{P}'_{b} \times \mathbf{P}'_{a}) \nonumber \\
  & = & - t'_{j}(\mathbf{P}_{b} - \mathbf{P}_{a})^\intercal ((\mathbf{R}_{\ell}\mathbf{v}_{\ell})\times \mathbf{f}'_{j})
        -{\mathbf{f}_{j}'}^\intercal (\mathbf{P}_{a} \times \mathbf{P}_{b}) \nonumber \\
  & = & 0 \nonumber \,. \hspace{6.7cm} \blacksquare
\end{eqnarray}

\subsection{Five-point Minimal Solver}
\label{sec:theory}

There exist six unknowns in our formulation. However, as mentioned in the above corollaries, there is an additional scale ambiguity in the line's support points. Hence, the inherent number of degrees of freedom is five, and a minimal solver for an eventail can be found by constructing five incidence constraints from five randomly picked events from the cluster. To remove the scale invariance, an additional constraint on the scale is added. Given that only the structure parameters are affected by the scale invariance, the scale constraint needs to include the related variables. We constrain the scale by adding the equation
\begin{equation}
  \left(\mathbf{R}_{\ell} \mathbf{v}_{\ell}\right)^\intercal \cdot \mathbf{R}_{\ell} \mathbf{v}_{\ell} - 1 = 0.
\end{equation}

Using Gr\"obner basis theory~\cite{cox2013ideals}, it is easy to find out that this problem has only two solutions. However, the fact that part of the solution variables is uniquely defined leads to some peculiarities in the derivation of the elimination template~\cite{kukelova2008polynomial}. Simply put, there are variable orderings for which the template leads to an equation where either $v_y$ or $v_z$ is a constant. More importantly though, after back substituting this variable into other equations, no simple solution for the remaining variables is apparent, but a second elimination template of substantial size needs to be solved.

To solve this problem, we analyzed all 720 possible monomial orderings and related elimination templates. We found out that 240 out of the 720 orderings directly lead to the unique solution for either $v_y$ and $v_z$ and are unlikely to reveal the double solution for the line's support points. The remaining 480 orderings and related elimination templates directly lead to the two solutions. The elimination templates can vary quite substantially in size and reach from $100\times 428$ to $244\times 820$.

\begin{figure}
  \centering
  \includegraphics[width=\columnwidth]{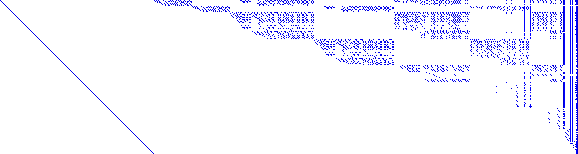}
  \caption{The chosen elimination template for our 5-point solver.}
  \label{fig:eliminationTemplate}
\end{figure}

The smallest template that directly leads to the two solutions is indicated in Figure \ref{fig:eliminationTemplate}, and has a size of $154\times 578$. Note that, given that the elimination leads to only two solutions, no action matrix decomposition is needed. In turn, the actual solutions are straightforward to recover from the last few rows of the template. Among these, we select the one resulting in the following variable ordering: $v_y > v_z > y_a > z_a > z_b > y_b$. In particular, it leads to one quadratic equation in $y_b$, which can be back substituted into three further equations to lead to univariate constraints in $y_a$, $z_a$, and $z_b$, respectively. With the structure parameters known, any two events can be used to construct bi-variate linear constraints on $v_y$ and $v_z$.

We embed the 5-point solver into RANSAC and use it to find the parameters of the eventail, which correspond to a partial dynamics recovery, as well as an event clustering. Sampling strategies and inlier criteria are discussed in Section~\ref{sec:experiments}.

\begin{figure*}[t]
    \centering
    \includegraphics[width=\linewidth]{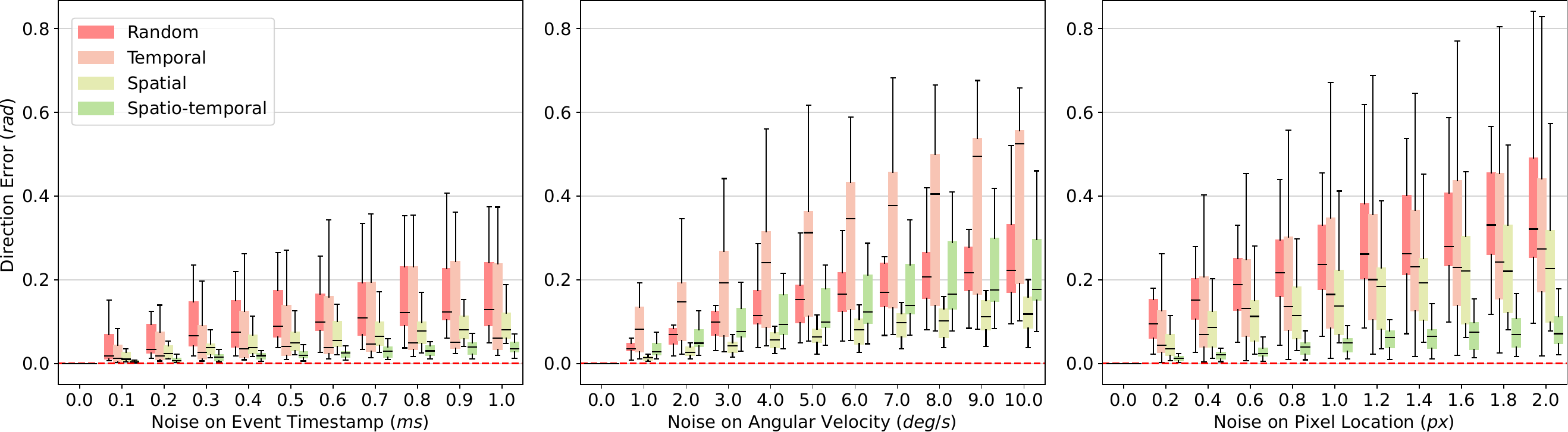}
    \caption{Results for the directional accuracy of the partially observed camera velocity as a function of noise in the event timestamps, the measured angular velocity, and the event locations. Each box denotes the range from the first quartile to the third quartile of the error distribution. The median is marked as the black line in the middle.}
    \label{fig:sim-a}
\end{figure*}

\subsection{Velocity averaging from multiple eventails}

We are now moving back to multiple eventails, which is why the index $i$ is reintroduced. The result we obtain from the clustering algorithm is a partial observation of the velocity $\mathbf{v}$ for each eventail. The observation is given by the two ratios $v\Line_{yi}$ and $v\Line_{zi}$ scaling the second and third basis vectors of $\mathbf{R}_{{\ell}i}$, respectively. Introducing the unobservable component $\kappa_i$ along the first basis vector, we have
\begin{equation}
  \mathbf{v} = \mathbf{e}\Line_{1i} \cdot \kappa_i + \mathbf{e}\Line_{2i} \cdot v\Line_{yi} + \mathbf{e}\Line_{3i} \cdot v\Line_{zi} \,.
\end{equation}
By multiplying the equation from the left with either the transpose of $\mathbf{e}\Line_{2i}$ or $\mathbf{e}\Line_{3i}$, and exploiting the orthogonal property of the basis vectors, we easily obtain
\begin{eqnarray}
  & & \left\{\begin{matrix}
    \mathbf{e}^{\ell\intercal}_{2i} \mathbf{v} = 
    \mathbf{e}^{\ell\intercal}_{2i}\mathbf{e}\Line_{1i} \cdot \kappa_i + 
    \mathbf{e}^{\ell\intercal}_{2i}\mathbf{e}\Line_{2i} \cdot v\Line_{yi} + 
    \mathbf{e}^{\ell\intercal}_{2i}\mathbf{e}\Line_{3i} \cdot v\Line_{zi} \\
    \mathbf{e}^{\ell\intercal}_{3i} \mathbf{v} = 
    \mathbf{e}^{\ell\intercal}_{3i}\mathbf{e}\Line_{1i} \cdot \kappa_i + 
    \mathbf{e}^{\ell\intercal}_{3i}\mathbf{e}\Line_{2i} \cdot v\Line_{yi} + 
    \mathbf{e}^{\ell\intercal}_{3i}\mathbf{e}\Line_{3i} \cdot v\Line_{zi}
  \end{matrix}\right. \nonumber \\
  & \Leftrightarrow & \left\{\begin{matrix}
    \mathbf{e}^{\ell\intercal}_{2i} \mathbf{v} = 
    \| \mathbf{e}^{\ell}_{2i} \|^2_2 \cdot v\Line_{yi} \\
    \mathbf{e}^{\ell\intercal}_{3i} \mathbf{v} = 
    \| \mathbf{e}^{\ell}_{3i} \|^2_2 \cdot v\Line_{zi}
  \end{matrix}\right. \Leftrightarrow \left\{\begin{matrix}
    \| \mathbf{e}^{\ell}_{2i} \|^{-2}_2 \cdot \mathbf{e}^{\ell\intercal}_{2i} \mathbf{v} = 
    v\Line_{yi} \\
    \| \mathbf{e}^{\ell}_{3i} \|^{-2}_2 \cdot \mathbf{e}^{\ell\intercal}_{3i} \mathbf{v} = 
    v\Line_{zi}
  \end{matrix}\right. \,. \nonumber
\end{eqnarray}
Stacking the result from all $N$ observed lines and taking into account that each individual partial velocity observation is affected by an unknown scale factor, we obtain the linear averaging scheme

\footnotesize
\begin{equation}
  \underbrace{\left[ \begin{matrix}
    \| \mathbf{e}^{\ell}_{21} \|^{-2}_2 \cdot \mathbf{e}^{\ell\intercal}_{21} & -v\Line_{y1} & \cdots & 0 \\
    \| \mathbf{e}^{\ell}_{31} \|^{-2}_2 \cdot \mathbf{e}^{\ell\intercal}_{31} & -v\Line_{z1} & \cdots & 0 \\
    \cdot & \cdot & \cdots & \cdot \\
    \cdot & \cdot & \cdots & \cdot\\
    \| \mathbf{e}^{\ell}_{2N} \|^{-2}_2 \cdot \mathbf{e}^{\ell\intercal}_{2N} & 0 & \cdots & -v\Line_{yN} \\
    \| \mathbf{e}^{\ell}_{3N} \|^{-2}_2 \cdot \mathbf{e}^{\ell\intercal}_{3N} & 0 & \cdots & -v\Line_{zN}
  \end{matrix} \right]}_{\left[\mathbf{A}\text{ }\mathbf{B}\right]}
  \left[ \begin{matrix} \mathbf{v} \\ \lambda_1 \\ \cdot \\ \cdot \\ \cdot \\ \lambda_N \end{matrix} \right] = \mathbf{0} \, ,
\end{equation}
\normalsize

\noindent where the additionally requested scaling factors are given by $\lambda_1, \ldots, \lambda_N$. Multiplying the equation from the left with $\left[\mathbf{A}\text{ }\mathbf{B}\right]^\intercal$, we obtain the form

\footnotesize
\begin{equation}
  \left[\begin{matrix}\mathbf{U} & \mathbf{W} \\ \mathbf{W}^\intercal & \mathbf{V}\end{matrix}\right]
  \left[\begin{matrix}\mathbf{v} \\ \lambda_1 \\ \text{:} \\ \lambda_N \end{matrix} \right] = \mathbf{0}\,,\text{ where}
\end{equation}
\normalsize

\footnotesize
\begin{eqnarray}
  \mathbf{U}_{3\times 3} & = & \sum_{i=1}^{N} \left(
    \frac{\mathbf{e}^{\ell}_{2i} \cdot \mathbf{e}^{\ell\intercal}_{2i}}{\|\mathbf{e}^{\ell}_{2i} \|^{4}_2} +
    \frac{\mathbf{e}^{\ell}_{3i} \cdot \mathbf{e}^{\ell\intercal}_{3i}}{\|\mathbf{e}^{\ell}_{3i} \|^{4}_2}
  \right) \nonumber \\
  \mathbf{V}_{N\times N} & = & \operatorname{diag}\left(
    (v_{y1}^{\ell})^2+(v_{z1}^{\ell})^2, \ldots, (v_{yN}^{\ell})^2+(v_{zN}^{\ell})^2
  \right)\nonumber \\
  \mathbf{W}_{N\times 3}^\intercal & = & \left[ \begin{matrix}
    -\frac{v_{y1}^{\ell}}{\| \mathbf{e}_{21}^{\ell} \|^2_2}\mathbf{e}_{21}^{\ell\intercal}-\frac{v_{z1}^{\ell}}{\| \mathbf{e}_{31}^{\ell} \|^2_2}\mathbf{e}_{31}^{\ell\intercal} \\
    \text{:}\\
    -\frac{v_{yN}^{\ell}}{\| \mathbf{e}_{2N}^{\ell} \|^2_2}\mathbf{e}_{2N}^{\ell\intercal}-\frac{v_{zN}^{\ell}}{\| \mathbf{e}_{3N}^{\ell} \|^2_2}\mathbf{e}_{3N}^{\ell\intercal}
  \end{matrix} \right]\nonumber \,.
\end{eqnarray}
\normalsize

\noindent Applying the Schur complement trick, we easily obtain
\begin{equation}
  \left[\mathbf{U}-\mathbf{W}\mathbf{V}^{-1}\mathbf{W}^\intercal\right]\mathbf{v}=\mathbf{0},
\end{equation}
which lets us find $\mathbf{v}$ via Eigen decomposition of a 3$\times$3 matrix. Note that $\mathbf{V}^{-1}$ is computed in linear time by simply inverting each element along the diagonal of $\mathbf{V}$.


\section{Experiments}
\label{sec:experiments}

We perform both simulation tests and real-world experiments. We first confirm the theoretical correctness of the proposed 5-point solver, its noise resilience, and discuss the implications of different event sampling strategies. Next, we test the influence of violations of the constant linear velocity motion assumption. We furthermore experimentally confirm the non-planar nature of the eventail manifold, and conclude with experiments on a public benchmark, demonstrating the advantage over existing bootstrapping methods. In order to evaluate the accuracy of the results, we adopt one of the criteria in~\cite{peng2021continuous}, which is the direction error $\phi$ between the estimated and the ground truth velocities, since the scale is not observable.

\subsection{Noise resilience on single cluster}

We start by evaluating the performance of the 5-point solver under different noise setups over synthetic data, and discuss the impact of different event sampling strategies. Samples from individual manifolds are generated as follows. We first randomly generate two lines in the image plane that represent the location of the line at the beginning and the end of the temporal window of events. The interval length is set to \SI{0.5}{\second}. Next, we sample randomly directed linear and angular velocities of \SI{1.0}{\meter/\second} and \SI{90}{\degree/\second} magnitude, respectively. By factoring in the interval duration, we can deduce the relative camera location between the beginning and the end of the interval by linear motion model, and extract the corresponding 3D line via triangulation. The line is furthermore given a finite length in 3D. The virtual event camera has a resolution of 640$\times$480 and a focal length of 320 pixels. The explained way of defining experiments ensures that the line passes through a sufficiently large area of the image canvas during the interval duration, thereby simulating well-posed problems where the camera exhibits sufficient displacement relative to the line in 3D. We randomly sample events, by first sampling a point on the 3D line, and then sampling the time stamp, at which that point is projected into the image plane. Each such projection is then denoted as an event. We analyze four strategies for sampling events in this way:
\begin{itemize}
    \item \textit{Random:} Event timestamps and 3D points on the line are both sampled randomly.
    \item \textit{Temporal:} the time interval is evenly divided into five sub-intervals, and each of the five events is assigned a random timestamp within an individual sub-interval.
    \item \textit{Spatial:} the line is evenly divided into five segments, and making sure that each of the five events is triggered by a random 3D point on an individual sub-segment.
    \item \textit{Spatio-temporal:} a combination of the spatial and temporal sampling strategies.
\end{itemize}

To conclude the experiment setup, we introduce three types of noise sources with different magnitudes, namely pixel noise, timestamp jitter, and noise to the camera's angular velocity, which is assumed to be given by an auxiliary sensor. The magnitude of the pixel noise and the noise on camera angular velocities is consistent within the same noise level but varies in direction. Zero-mean Gaussian noise is used for timestamp noise. Results are presented in Figure~\ref{fig:sim-a}. Each box in the plot represents the mean errors from fifteen different geometry-motion configurations. Within each configuration, we conducted 100 evaluations with different event samples. Note that since the 5-point solver only takes readings from one cluster and produces partial observation, we compare the estimated velocity with the normalized ground truth velocity in the direction perpendicular to the line. Moreover, as is standard in SLAM evaluation pipelines, we do not report evaluations of the line parameters, as their error is subsumed in the motion errors.

Errors generally increase as noise levels increase. Without noise, our solver can always produce zero-error results, hence proving the theoretical correctness of the proposed method. There does not exist, however, a golden sampling strategy to produce minimum errors for different noise sources, though ensuring spatial distribution among the five sampled events is crucial to achieve accurate results. In the following RANSAC experiments, we therefore alternate between spatial and spatio-temporal sampling.

\subsection{Multiple clusters and validity of motion model}

We have developed an Event-based Geometric wireframe Generator, entitled EGG, in order to conduct experiments over longer time periods. The wireframes are composed of finite lines which simulate strong gradient locations in the scene. Note that the direction of the gradient does not need to be specified, given that we ignore polarity in our formulation. The simulator supports different types of motion models, including both spline-based and non-polynomial motion. Events are generated whenever a projected line comes across a pixel center, thereby producing events with accurate timestamps. A realistic IMU model is also implemented with bias and random walk. In a nutshell, the EGG simulator we have developed is capable of producing accurate ground truth readings of poses and twists, corrupted IMU readings, as well as line-generated events with precise timestamps.

We now investigate the robustness and accuracy of the overall velocity determination from multiple eventails. We generate ten line segments within the volume $[-2, 2]\times[-2, 2]\times[1.5, 3.0]$. Each event is labeled with its corresponding line to bypass clustering in the present experiment. In the next two sections, we evaluate the full RANSAC pipeline. Note, however, that we are still running RANSAC to fit the eventail parameters to each event cluster. We adopt the angular reprojection error~\cite{lee2019closed} within the RANSAC algorithm. We conduct two types of motion, both of which violate the constant linear motion assumption. The first moves in a circular arc, with a tangential velocity of \SI{1.0}{\meter/\second} and with a radius ranging from \SI{2}{\m} to \SI{10}{\m} in increments of \SI{2}{\m}. Each sequence lasts \SI{0.3}{\second}, and so does the chosen time window. For each configuration, we generate twenty sets of events without noise and another twenty with noise. Perturbations are given by zero-mean Gaussian noise with a standard deviation of 1 pixel on location and \SI{1}{\ms} on timestamp. We also use corrupted IMU readings with a realistic bias. Finally, the ground truth camera motion has no rotation and faces in a constant direction. In the second simulation, we add acceleration with magnitude ranging from \SI{0.1}{\m/\second\squared} to \SI{0.5}{\m/\second\squared} to the linear velocity. Again, the time window is \SI{0.3}{\second}.

\begin{figure}
    \centering
    \includegraphics[width=\linewidth]{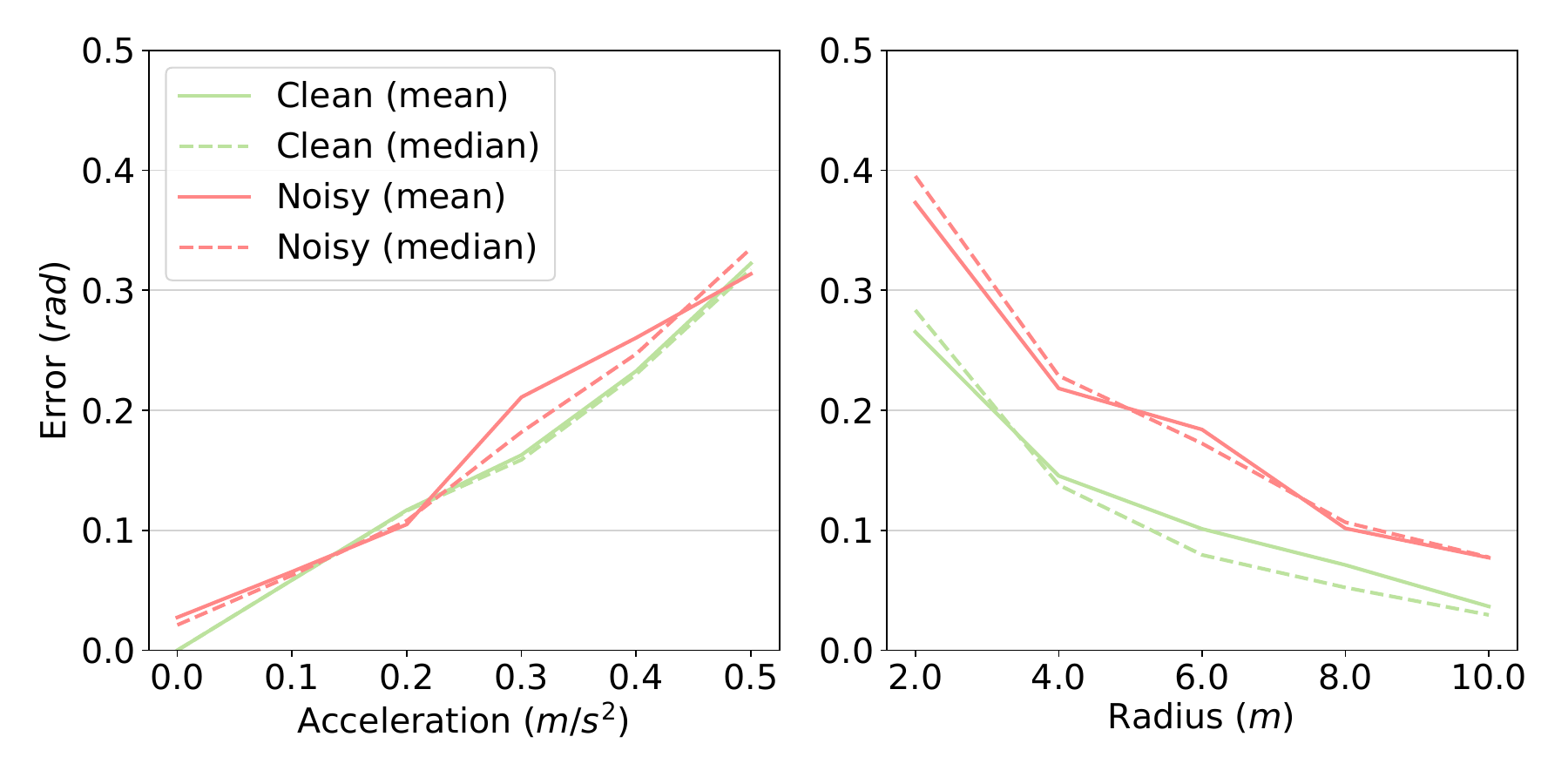}
    \caption{Average directional errors of fully estimated linear velocity over ten \'eventails. Results are evaluated for clean and noisy data, and different violations of the motion model assumptions.}
    \label{fig:sim-b}
\end{figure}

Figure~\ref{fig:sim-b} shows that without noise and model violations, our method can always produce zero-error results. As the velocity becomes less constant (larger acceleration, smaller radius), the error increases, but the proposed solver still shows high noise robustness in these conditions. 

\subsection{Performance under high dynamics}
\label{sec:dna}

To demonstrate the full potential of our parametrization, we generate sequences of \SI{1}{\second} with constant but very significant linear and angular velocities. Two non-parallel lines with comparable depth, \ie from $[0, 0.75, 3]$ to $[0, 2, 3]$ and from $[0.38, -0.65, 3]$ to $[0.75, -1.3, 3]$ respectively, are placed in front of the camera. The camera is moving towards the lines with a linear velocity of $[0.4, 0.4, 2.0]$\SI{}{{\meter/\second}}, and with a self-rotation of $[0, 0, -2\pi]$\SI{}{{\radian/\second}}. The noise level is kept the same as in the previous simulation. As showcased in the example in Figure~\ref{fig:front}, our method successfully fits the two manifolds from five events with inlier ratios of \SI{67.16}{\percent} and \SI{73.22}{\percent}, respectively, and a velocity direction error of only \SI{0.01}{\radian}. On the other hand, traditional plane-based fitting fails in this case and splits up the set into 34 subsets. As explained next, this difference in clustering will impact the achievable final accuracy.

\subsection{Real-world Experiment}

We validate the method on four sequences from a public benchmark dataset~\cite{gao2022vector} which feature clear line structures. The dataset provides VGA event recordings from a Gen3 Prophesee camera, \SI{200}{\Hz} ground truth camera poses from a MoCap system, and \SI{200}{\Hz} measurements from an XSens MTi-30 AHRS IMU. To increase efficiency we downsample the events in each sequence by a factor of ten. We then split the sequence into non-overlapping intervals of events with a \SI{0.3}{\second} duration each. To find individual line clusters we adopt two different approaches: In the first approach, we find clusters using cilantro~\cite{zampogiannis2018cilantro} which finds spatio-temporal planes, as was used in~\cite{peng2021continuous}, then when sampling 5 tuples of events from these clusters we count inliers over the entire event set, which has higher robustness than simply considering events within that cluster, as was done in~\cite{peng2021continuous}. In this experiment we set the maximum number of event clusters to five. The second one is unique to our method, and consists of running RANSAC on the entire window of events until a maximal inlier ratio is found, then extracting that line and partial velocity estimate together with the inlier events from the event set, and repeating this step for a total of five times. We use this strategy whenever cilantro fails to provide sufficient clusters. This means that our method is strictly more robust than~\cite{peng2021continuous} which cannot recover when cilantro fails. The estimated velocity is obtained by velocity averaging from multiple eventails and is further compared with the ground truth velocity in direction error $\phi$. Comparative results are listed in Table~\ref{tab:real-world} reporting the mean and median error of successful samples in each sequence. Since CELC+opt fails on a subset of the sequences, we also restrict the evaluation of our method to the subsequence where CELC+opt is successful, and denote these results with an asterisk $^*$. Where there are too few clusters, CELC+opt fails to output valid results in some samples.\\ 
\textbf{Results} We find that while the success rate of~\cite{peng2021continuous} ranges between 23\% and 70\% our method consistently achieves 100\%. Moreover, on the subset where~\cite{peng2021continuous} is successful both methods are comparable in terms of accuracy. This highlights that our methods signficantly improves on the robustness of existing closed-form solvers for linear velocity. 

\begin{table*}[t]
  \caption{Real-world results on the VECtor benchmark~\cite{gao2022vector}. We report the success rate, \ie, the percentage of sequence's sections where the algorithm outputs reasonable results, as well as the velocity direction error $\phi^*$ only computed in these sub-sections. For our method, which always has \SI{100}{\percent} success rate, we also report the error $\phi$ over the full sequence.}
  \label{tab:real-world}
  \begin{center}
  \resizebox{0.75\textwidth}{!}{%
  \begin{tabular}{c|ccc|cccccc}
    \toprule
    \multirow{2}{*}{Seq. Name} & \multicolumn{3}{c|}{CELC+opt~\cite{peng2021continuous}} & \multicolumn{5}{c}{Ours} \\ 
    & $\phi^*_\text{mean}$ & $\phi^*_\text{median}$ & Success & $\phi^*_\text{mean}$ & $\phi^*_\text{median}$ & $\phi_\text{mean}$ & $\phi_\text{median}$ & Success\\ 
    \midrule
    \emph{board-slow}      & 0.451 & 0.434 & \SI{65.69}{\percent} & \textbf{0.429} & \textbf{0.385} & 0.484 & 0.416 & \textbf{\SI{100}{\percent}} \\
    \emph{mountain-normal} & \textbf{0.483} & \textbf{0.512} & \SI{56.70}{\percent} & 0.542 & 0.528 & 0.584 & 0.586 & \textbf{\SI{100}{\percent}} \\
    \emph{desk-normal}     & 0.464 & \textbf{0.464} & \SI{69.86}{\percent} & \textbf{0.461} & 0.474 & 0.461 & 0.466 & \textbf{\SI{100}{\percent}} \\
    \emph{sofa-normal}     & \textbf{0.419} & 0.455 & \SI{23.16}{\percent} & 0.532 & \textbf{0.438} & 0.550 & 0.514 & \SI{100}{\percent} \\
    \bottomrule
    \multicolumn{5}{l}{$^*$ subset where CELC+opt~\cite{peng2021continuous} does not fail} \\\hline
  \end{tabular}
  }    \vspace{-4ex}
  \end{center}
\end{table*}


\section{Conclusion and Future Work}

The present work provides a new understanding of the geometry of line-generated events perceived under constant linear and angular velocities. We have derived a theoretically correct parametrization of the manifold-distribution of such events, and showed how the parametrization partially involves camera dynamics parameters. Our presented minimal solver for the manifold parameters has been successfully embedded into RANSAC, based on which we studied resilience against noise and motion model violations. Extensive tests on real-world data have validated the superior ability of our method to fit \'eventail manifold parameters, thereby increasing accuracy and the overall success rate of the presented bootstrapping algorithm. 

Prospective future directions include extending the proposed averaging scheme with uncertainty estimates from multiple manifolds, as well as optimizing over the cluster inliers for improved robustness and accuracy of the results. Moreover, while we showed that our sequential RANSAC approach was sufficient for fitting multiple lines into a given event stream, more sophisticated robust multi-model fitting techniques like Progressive-X~\cite{Barath19iccv} could be used to minimize the number of outliers. Finally, the current method omits the event polarity and does not perform any temporal smoothing or fusion of accelerometer or gyroscope readings from the IMU. Extending the approach in this direction would make it acceleration-aware, as~\cite{chamorro2020high}, and lead to improved modeling of the camera motion. Despite these limitations, we believe that our findings lay the cornerstones for highly successful, incremental smoothing-based motion estimation.


\section*{Acknowledgments}

\noindent The research presented in this paper has been supported by projects 22DZ1201900 and 22ZR1441300 funded by the Shanghai Science Foundation as well as project 62250610225 by the National Science Foundation of China. This work was also supported by the Swiss National Science Foundation and the European Research Council (ERC) under grant agreement No. 864042 (AGILEFLIGHT). 


{\small
\bibliographystyle{ieee_fullname}
\bibliography{egbib}
}

\end{document}